\documentclass[10pt,twocolumn,letterpaper]{article}

\usepackage[pagenumbers]{cvpr} 

\usepackage{graphicx}
\usepackage{amsmath}
\usepackage{amssymb}
\usepackage{booktabs}
\usepackage{bbm}

\usepackage[ruled]{algorithm2e}

\usepackage{textcomp}
\usepackage{siunitx}

\usepackage{xcolor}

\usepackage[pagebackref,breaklinks,colorlinks]{hyperref}

\usepackage[capitalize]{cleveref}
\crefname{section}{Sec.}{Secs.}
\Crefname{section}{Section}{Sections}
\Crefname{table}{Table}{Tables}
\crefname{table}{Tab.}{Tabs.}

\DeclareMathOperator*{\argmin}{arg\,min}

\newcommand{\X}{\mathbf{X}}
\newcommand{\Y}{\mathbf{Y}}

\newcommand{\w}{\mathbf{w}}
\newcommand{\m}{\mathbf{m}}
\newcommand{\e}{\mathbf{e}}

\def\cvprPaperID{5592} 
\def\confName{CVPR}
\def\confYear{2022}

\newcommand\blfootnote[1]{%
  \begingroup
  \renewcommand\thefootnote{}\footnote{#1}%
  \addtocounter{footnote}{-1}%
  \endgroup
}

\begin{document}

\title{The Two Dimensions of Worst-case Training \\
and Their Integrated Effect for Out-of-domain Generalization}

\author{
Zeyi Huang$^{1,2\star}$ \hspace{0.6cm} Haohan Wang$^{1\star}$ \hspace{0.6cm} Dong Huang$^1$ \hspace{0.6cm} Yong Jae Lee$^{2\dagger}$ \hspace{0.6cm} Eric P. Xing$^{1\dagger}$ 
\\
{$^1$Carnegie Mellon University \hspace{0.6cm} $^2$University of Wisconsin-Madison}\\
\tt\small{ \{zeyih@andrew, haohanw@cs, donghuang@, epxing@cs\}}.cmu.edu \hspace{0.3cm} \tt\small{ yongjaelee@cs.wisc.edu}\\
}
\maketitle

\begin{abstract}
Training with an emphasis on ``hard-to-learn'' components of the data 
has been proven as an effective method 
to improve the generalization of machine learning models,
especially in the settings where robustness (e.g., generalization across distributions) is valued. 
Existing literature discussing this ``hard-to-learn'' concept are mainly expanded 
either along the dimension of the samples or the dimension of the features. 
In this paper, we aim to introduce a simple view merging these two dimensions, 
leading to a new, simple yet effective, heuristic to train machine learning models 
by emphasizing the worst-cases on both the sample and the feature dimensions. 
We name our method W2D following the concept of ``Worst-case along Two Dimensions''. 
We validate the idea and demonstrate its empirical strength over standard benchmarks. 
\end{abstract}

\section{Introduction}
\blfootnote{$^\star$,$^\dagger$ equal contribution. Code is avaiable at \href{https://github.com/OoDBag/W2D}{\color{black}{github.com/OoDBag/W2D}}}
\begin{figure}
    \centering
    \includegraphics[width=0.45\textwidth]{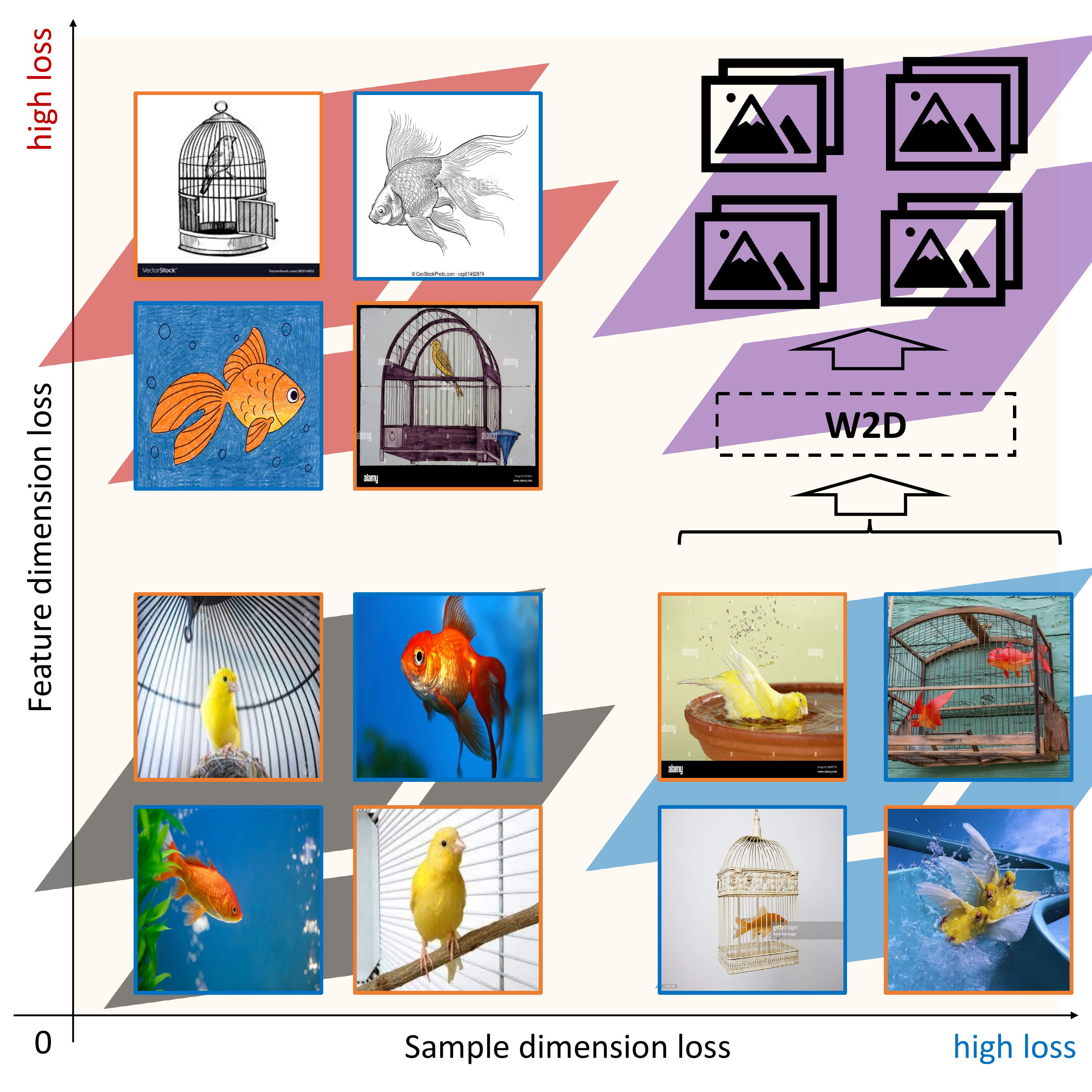}
    \caption{The conceptual illustration of our main idea W2D for a simple example of canary vs. goldfish image classification. 
    For a regular model trained on standard images (left-bottom block),
    there are two dimensions of hard samples:
    following the categorization in \cite{ye2021ood}, 
    the vertical dimension 
    corresponds to the ``diversity shift'' of the images 
    (\textit{e.g.}, photos of the animals vs. cartoons of the animals)
    and the horizontal dimension 
    corresponds to the ``correlation shift'' of the images 
    (\textit{e.g.}, birds in cage and fish in water vs. fish in cage and birds in water). 
    The W2D algorithm conceptually selects the images that are hard at the sample dimension and augment these samples toward being harder at the feature dimension. 
    }
    \label{fig:main}
\end{figure}

The remarkable empirical performance of deep learning over 
\textit{i.i.d} data, 
sometimes paralleling the human visual system \cite{krizhevsky2012imagenet,he2015delving}, 
has encouraged the community to challenge 
potentially more demanding scenarios 
where the models are trained with 
data from one or more distributions 
but tested with data from other distributions. 
We refer to this scenario as the out-of-distribution (OOD) generalization
testing setting following the terminology used in \cite{ye2021ood}. 

In this OOD test scenario, 
deep learning techniques often underdeliver the promising results made with \textit{i.i.d} data, 
as observed by multiple preceding works
with different strategies to generate the test data, such as
with salient patterns added to the data \cite{geirhos2018imagenettrained,hendrycks2018benchmarking}, 
with carefully constructed imperceptible noise perturbing the data (adversarial attacks) \cite{SzegedyZSBEGF13,GoodfellowSS14}, 
or with additionally collected datasets 
that humans can nonetheless generalize to despite potentially significant disparities between the training and test distributions (\textit{e.g.}, domain adaptation/generalization) \cite{ben2010theory,muandet2013domain}.

Despite the variation in multiple OOD settings, 
the underlying reason leading to the performance drop may have a shared theme:
the models' incapability to learn what humans will consider important in the data,
as discussed previously in empirical \cite{wang2020high} and statistical \cite{wang2021toward} perspectives. 

Thus, we conjecture that a key to training models 
that can perform consistently well in the OOD setting is 
to design a new training heuristic that can 
better imitate the human's learning behaviors. 
In addition, we also hope the new heuristic is 
simple and general so that it
can be directly plugged into 
and benefit existing methods 
across different architectures, optimizers, losses, or regularizations. 

\emph{A psychological prior:} 
In seeking the answer of how a human can learn most efficiently, 
we notice that a world-renowned psychologist 
(Dr.~K.~Anders Ericsson)
has devoted his life-long career 
decoding the habits of people with expert-level performances. 
His main conclusion \cite{ericsson1993role} is that the high-end performances are the result of extensive practices 
\emph{beyond one's comfort zone}. 

Back to the discussion of machine learning, 
we analogize the \emph{beyond comfort zone} elements of one's daily life
to the elements of the training data that are particularly \emph{hard to learn} for a model.
We notice this ``element'' can be interpreted with two perspectives:
one interpretation is that a certain pattern across many images is hard to learn; 
and the other is that some specific images in the dataset are hard to learn.  

Previous discussions devoted exclusively 
to either one of these two elements 
has been expanded extensively. 
For example, 
a line of methods have been invented 
to counter the model's tendency to learn some simple patterns \cite{WangHLX19,bahng2019learning,WangGLX19,nam2020learning}, 
and another line of methods 
have been introduced 
to push the model to learn 
patterns represented by a small set of samples \cite{sagawa2019distributionally,krueger2021out}. 
However, 
there seems to be no discussion aiming to 
train the model to overcome the limitations raised from both of these perspectives, 
while doing so would intuitively improve the model's performances, 
as well as 
align well with the psychological findings mentioned above. 

In this paper, 
inspired by the psychological prior above, 
we aim to introduce a simple training heuristic 
that will push the model to learn 
the hard-to-learn concept 
on both the feature dimension and the sample dimension. 
As our method is intuitively a combination 
of worst-case training at the feature level 
and worst-case training at the sample level, 
the new technique can serve as a simple heuristic to replace 
the existing training procedure of deep learning models 
regardless of model architecture, optimizer, loss, or regularization \textit{etc}, 
as long as the optimization is within the gradient descent family. 
We name our method \textsf{W2D} following the concept of ``Worst-case along Two Dimensions''.

The remainder of this paper is organized as follows. 
In Section~\ref{sec:related},
we first introduce the background of this paper, 
with an emphasis on the ``worst-case training'' along the two dimensions, 
and their corresponding effects on OOD generalization, 
which inspired us to investigate the integrated effect of these two worst-case training dimensions. 
In Section~\ref{sec:method}, 
we introduce our new heuristic that combines these two directions, and we demonstrate the method's empirical strength in Sections~\ref{sec:exp} and~\ref{sec:ablation}. 
We offer several related discussion in Section~\ref{sec:dis} before we conclude with Section~\ref{sec:con}. 

\section{Background}
\label{sec:related}

We introduce the background of our work in this section. 
We first offer a brief summary of works that 
improve a model's OOD generalization performance.  We  
then focus on related work that are 
devoted to solving the two challenges
at the feature level and at the sample level, respectively. 
As we notice that some of the methods solving these two problems 
have a common theme of emphasizing the hard-to-learn elements from the data, 
we continue the discussion with a focus on the worst-case training methods
along both dimensions. 
Finally, we wrap up this section 
with a summary of the key contributions made in this paper.

\subsection{Domain Adaptation, Domain Generalization, and New Paradigms}

The investigation of a model's generalization ability across distributions 
can probably be traced back to the study 
of domain adaptation \cite{ben2007analysis,ben2010theory}, 
which studies the general problem of maintaining  
a model's performance over a test distribution that is different 
from the training distribution. 
Early-stage theoretical work suggests 
one of the key factors of learning cross-domain generalizable models 
is to enforce invariance across distributions \cite{ben2010theory},
and this has inspired a long line of work aiming to learn invariant representations across the training and testing distributions \cite{MansourMR09,saenko2010da,GermainHLM16,ZhangLLJ19,dhouib2020margin,wang2020dictionary,fan2020domain,kim2020learning}, 
with the most popular recent examples 
being domain adversarial neural networks \cite{ganin2016domain}. 

Domain generalization \cite{muandet2013domain} is another mainstream research topic 
in the study of OOD generalization. 
It extends the setup of domain adaptation to a setting 
in which the testing distribution data, even unlabelled, is not available during training. 
Instead, models are trained with data from multiple training distributions, 
and enforcing invariance across these training distributions has become a major 
theme \cite{ghifary2015domain,wang2016select,motiian2017unified,li2018domain,carlucci2018agnostic,akuzawa2019adversarial,nguyen2021domain,rahman2021discriminative,han2021learning}. 

However, recently, the paradigm of learning invariant representations 
has been challenged by the argument that 
invariance is not sufficient for cross-domain 
generalization if the data have different labelling functions \cite{wu2019domain,ZCZG19}, 
potentially leading to a new paradigm of learning 
across distributions with disparate labelling functions \cite{wang2021toward}, 
as will be detailed with examples in the next section.


As a result, recent studies are not always bounded by the concepts of 
domain adaptation or domain generalization, 
but are conducted in the new paradigm with disparate labelling functions.

\subsection{Hard-to-learn Patterns and Solutions}

One of the mainstream studies in robust machine learning 
focuses on the issue of 
models learning some patterns in the training data that 
are not present in the test data, 
with the most popular example probably being the 
snow background in husky vs. wolf classification \cite{ribeiro2016should}. 
This problem is often referred to as biases \cite{torralba2011unbiased}, spurious features \cite{vigen2015spurious}, confounding factors \cite{mcdonald2014confounding}, 
or superficial features \cite{WangHLX19}, 
but the solution to counter the problem usually has
a unified theme of leveraging the 
human knowledge of the differences between training and testing distributions
to either regularize the hypothesis space \cite{WangHLX19,bahng2019learning,WangGLX19,nam2020learning} or to augment the data \cite{geirhos2018imagenettrained,hermann2020origins,wang2020squared,hendrycks2019augmix}, 
as summarized in \cite{wang2021toward}. 

Interestingly, the RSC method \cite{huang2020self} also aims to solve the challenge along this line, 
but it does not require prior knowledge over the patterns. 
Building upon an assumption that 
learning all the features,
instead of just the most discriminative ones,
will benefit the OOD generalization, 
RSC essentially uses a selective dropout mechanism 
to perform augmentation, 
and achieves good benchmark performances 
on popular OOD datasets \cite{huang2020self,ye2021ood}.
Conceptually, RSC prepares the features for each sample
by dropping out the most predictive features 
(\textit{i.e.}, creating features that are challenging for the model to learn).


\subsection{Hard-to-learn Samples and Solutions}

On the other hand, hard-to-learn samples pose a different challenge:
some samples in the training data are ignored by the model 
because these samples are considered to be the ``minority'' in the training set \cite{mehrabi2021survey,chouldechova2020snapshot}. 
To counter this problem, 
training procedures that emphasize the minority samples have been introduced, such as the family of DRO methods \cite{NIPS2016_4588e674,hu2018does,sagawa2019distributionally,NEURIPS2020_07fc15c9,nam2020learning}, 
with different strategies to identify the minority samples 
and to interpolate the training set according to weighting factors that favor the minority samples. 
Intuitively, these methods prepare the batches of samples with an emphasis 
on the samples that are challenging for the model to learn. 

Further along this line, the community extends the idea of interpolation to extrapolation by adjusting the weighting factors so that the weights for simpler samples can even be negative, to further push the models to focus on the hard-to-learn samples. 
The VREx method \cite{krueger2021out} is introduced in this context 
and achieves leading empirical performances on benchmarks \cite{krueger2021out,ye2021ood}. 

\subsection{Worst-case Training in Each Dimension and the Corresponding Effects}

With the proliferation of methods introduced to improve models' OOD performances, 
and various empirical claims, there have been some integrated comparisons of the various methods. 
For example, 
interestingly, 
DomainBed suggests that all these new methods 
are still shy of the conventional Empirical Risk Minimization (ERM) method \cite{gulrajani2020search}
under their extensive range of hyperparameter choices. 
While this message sets a striking alarm to the community, 
it may seem overly pessimistic (more details on this are offered later in Section~\ref{sec:dis}).


Recently, the OOD-bench \cite{ye2021ood} extends the spirit of DomainBed 
but offers a more finer-grained analysis of the models' performances. 
It comprehensively examines the performances of the recent models 
over popular benchmark datasets, 
but with separate discussions 
on ``diversity shift`` datasets and ``correlation shift'' datasets.
Diversity shift datasets refer to the benchmarks with a relatively significant style shift from training distribution to the test distribution (such as from photos to sketches), 
and correlation shift datasets refer to the benchmarks with clear defined spurious features
that are correlated with the label 
(such as when the color of the digits are associated with the labels of the digits in the image digit classification task).


OOD-bench investigates the popular methods along these two directions, and shows that for each direction, there are only a couple of methods that outperform ERM. 
Interestingly, the best performing method for diversity shift is RSC \cite{huang2020self}, 
one of the most advanced methods aimed to learn hard-to-learn patterns, 
with a heuristic to push the model to train with generated worst-case features.  On the other hand, the best performing method for correlation shift is VREx \cite{krueger2021out}, 
one of the most advanced methods aimed to learn hard-to-learn samples,
with a heuristic to push the model to train with selected worst-case samples.

Our contribution is inspired by the above discussion: 
if worst-case training in the feature dimension excels at diversity shift,
while worst-case training in the sample dimension excels at correlation shift, 
if we can integrate these two worst-case training methods into one simple heuristic, 
the new method will likely lead to sufficiently good empirical performance for both diversity and correlation shift.

\subsection{Key Contributions}
In comparison to the previous methods discussed along these two lines, 
we believe the key contributions we make in this paper are as follows. 
\begin{itemize}
    \item We discuss the previous methods with a unified theme of worst-case training along two different dimensions, and doing so naturally leads to the integration of these methods.
    \item We introduce a new method, called \textsf{W2D}, as an integration of these two types of training methods. 
    W2D is a simple heuristic that can be directly plugged into any training process regardless of model architecture, loss function, regularization or optimizer. 
    \item 
    We demonstrate strong empirical performance on multiple benchmark datasets, 
    and conduct ablation studies to understand the contribution of each of component of the method. 
\end{itemize}

\section{Method}
\label{sec:method}
In this section, we first formalize the two worst-case training methods, 
which naturally leads to the introduction of our method. 
Then, with the main framework of our method introduced, 
we continue to discuss a whole-batch patching heuristic we use in the experiments 
that have benefited our method empirically with non-negligible margins.

\subsection{W2D Method}
We first introduce our notations. We use $(\X, \Y)$ to denote a dataset with $n$ (data,label) paired samples. 
Thus $\X \in \mathcal{R}^{n \times p}$ and $\Y \in \mathcal{R}^{n}$. 
We use $f(\cdot; \theta)$ to denote the model we aim to train, 
and 
use $e(\cdot; \theta_e)$ and $d(\cdot; \theta_d)$ to denote encoder and decoder, respectively. 
Thus, 
we have $f(\cdot; \theta) = d(e(\cdot;\theta_e); \theta_d)$. 
We use $\w$ to denote a weight vector of length $n$. 
We use $\m$ to denote a masking vector with some elements to be 0 and others to be 1; the length of $\m$ is the same as the feature dimension (the output of $e(\cdot; \theta_e)$). 
We use $l(\cdot, \cdot)$ to denote a generic loss function. 

The vanilla training process of a model is
\begin{align*}
    \widehat{\theta}_{\textnormal{vanilla}} = \argmin_\theta \dfrac{1}{n}\sum_i l(f(\X_i;\theta), \Y_i)
\end{align*}

\paragraph{Worst-case along feature dimension}
We formalize the first worst-case method, with a generic form as follows: 
\begin{align}
    \widehat{\theta}_{\textnormal{w\_feature}} = 
    \argmin_\theta \dfrac{1}{n}\sum_i \max_\m l(d(\m \odot e(\X_i;\theta_e);\theta_d), \Y_i), 
    \label{eq:rsc}
\end{align}
where $\odot$ denotes the element-wise product. 

In particular, the RSC method \cite{huang2020self} introduces the $\m$ with a hyperparameter $\rho$, which denotes the $\rho$ fraction of the elements are zeros. 
The maximization step is achieved by examining the magnitude of the gradient of $\frac{\partial d(\e;\theta_d)}{\partial \e}$. 

\paragraph{Worst-case along sample dimension}
We formalize the second worst-case method, with a generic form as follows:
\begin{align*}
    \widehat{\theta}_{\textnormal{w\_sample}} =& 
    \argmin_\theta \dfrac{1}{n}\sum_i \max_{\w_i} \w_i l(f(\X_i;\theta), \Y_i), \\
    & \textnormal{subject to} \sum_i \w_i=1
\end{align*}
In general, the choice of $\w_i$ depends on $l(f(\X_i;\theta), \Y_i)$, with concrete differences across different methods such as \cite{sagawa2019distributionally,NEURIPS2020_07fc15c9,nam2020learning}.
A common theme is that, the higher the loss is, the bigger $\w_i$ is. 

In practice, because we use batch-wise optimization, 
the estimation of $\w$ is not straightforward. 
Fortunately, we can use a simple alternative: 
for each batch, 
we select the samples with high losses (through a forward pass), 
and then use these samples to update the model. 
This is a heuristic used by multiple methods such as \cite{chang2017active,xue2019hard,katharopoulos2018not,byrd2019effect}.

\paragraph{W2D Method}
Integrating the two methods above, 
we have
\begin{align*}
    \widehat{\theta}_{\textnormal{W2D}} = &
    \argmin_\theta \dfrac{1}{n}\sum_i \max_{\m, \w_i} \w_i l(d(\m \odot e(\X_i;\theta_e);\theta_d), \Y_i), \\
    & \textnormal{subject to} \sum_i \w_i=1
\end{align*}

In practice, 
we use the RSC method to identify the $\m$ for worst-case training in the feature dimension, and use the above heuristic to perform worst-case training in the sample dimension.

\begin{algorithm}[t!]
\SetAlgoLined
 \textbf{Input:} data set $(\X, \Y)$, percentage of samples used per batch $\rho$, percentage of whole batch patching $\kappa$, batch size $\eta$, maximum number of epochs $T$, and other RSC hyperparameters\;
 \textbf{Output:} Classifier $f(\cdot;\theta)$\;
 randomly initialize the model $\theta_0$\;
 calculate the number of iterations $K = n/\eta$\;
 \While{$t \leq (1-\kappa)T$}{
    \For{a batch of data $(\X, \Y)_k$ where $k \leq K$}{
        forward pass to calculate the loss $l(f(\X_i;\theta_{t, k-1}),\Y_i)$ of every sample in the batch\;
        select the top $\eta\rho$ samples with highest loss to construct $(\X, \Y)_{k, \rho}$\;
        Train the model with $(\X, \Y)_{k, \rho}$ following \eqref{eq:rsc}.
        
    }
 }
 \While{$(1-\kappa)T < t \leq T$}{
    \For{a batch of data $(\X, \Y)_k$ where $k \leq K$}{
        Train the model with $(\X, \Y)_k$ following \eqref{eq:rsc}.
    }
 }
 \caption{W2D Algorithm}
 \label{alg:main}
\end{algorithm}


\subsection{Whole-batch Patching Heuristic} \label{sec:trainingheuristics}

As introduced above, W2D selects the worst-case training samples with highest loss in each training iteration, as the model evolves over training, 
it is possible that what was once considered easy becomes hard, and vice versa. 
Therefore, we can intuitively expect the 
model to see all the samples in the training set given sufficient training iterations. 
However, chances are that some of the samples are never seen by the model during training as these samples are always considered not hard enough;
it is obvious not ideal that there is a chance the model does not take full advantage of the training set. 

To counter this potential issue, we simply switch to whole batch training during the last $\kappa\%$ of training epochs, which leads to better empirical performances across different model selection strategies.
We verify the effectiveness of this simple approach in our ablation study. 
More results can be found in Section \ref{sec:trainvstest}.

The full description of W2D with the whole batch patching heuristic is detailed in Algorithm~\ref{alg:main}. 

\section{Experiments}
\label{sec:exp}
\subsection{Experimental Setup}

We follow the setting in \cite{ye2021ood} and evaluate domain generalization on both types of distribution shift: diversity shift and correlation shift. Specifically, we use the same strategy for model selection, dataset splitting, and network backbone. More details of the experimental settings can be found in the discussion and supplementary materials. 

\subsection{Datasets, Hyperparameter Search and Model Selection}
We follow the Ood-bench~\cite{ye2021ood} to choose datasets that cover as much variety as possible from the various OoD research areas for our experiments. We conduct experiments on seven OOD datasets: CMNIST~\cite{arjovsky2019invariant}, CelebA~\cite{liu2015deep}, NICO~\cite{he2021towards}, Terra Incognita~\cite{beery2018recognition}, OfficeHome~\cite{venkateswara2017deep}, WILDS-Camelyon,~\cite{koh2021wilds} and PACS~\cite{li2017deeper}.
These datasets are divided into two categories based on their estimated diversity and correlation shift.

We use the same hyperparameter search protocol as \cite{ye2021ood, gulrajani2020search}: a 20-trials random hyperparameter search is conducted for every dataset and algorithm pair. The random search study is repeated for total three series of hyperparameter combinations, weight initializations, and dataset splits. The three series yield the three best accuracies in total over which a mean and standard error bar is computed for every dataset-algorithm pair.

To be consistent with existing line of work, models trained on PACS, OfficeHome, and Terra Incognita are selected by training-domain validation;
models trained on WILDS-Camelyon and NICO are selected by leave-one-domain-out validation;
while models trained on Colored MNIST and CelebA are selected by test-domain validation. 
Details of these selection strategies can be found in \cite{ye2021ood}.


\subsection{Empirical Results}

The benchmark results are shown in Table \ref{table:diversityshift} and Table \ref{table:correlationshift}. In addition to mean accuracy and standard error bar, we follow the Ood-bench~\cite{ye2021ood} to report a ranking score for each algorithm with respect to Empirical Risk Minimization (ERM)~\cite{vapnik1999nature}. Specifically, depending on whether the attained accuracy is lower than, within, or higher than the standard error bar of ERM accuracy on the same dataset, scores -1, 0, +1 are assigned to every dataset-algorithm pair. The ranking score for each algorithm is computed by adding up the scores across all datasets listed in the table. The ranking score reflects a relative degree of robustness against diversity and correlation shift compared to ERM.


Note, for CMNIST, Ood-bench \cite{ye2021ood} evaluates the results using the -90 as testing domain while DomainBed~\cite{gulrajani2020search} reports the results averaged over the +90, +80, and -90 domains. We follow Ood-bench's setting to report the results in Table~\ref{table:correlationshift} and we also report the results of the DomainBed's setting in the supplementary material. 
The choice of the settings does not affects our ranking score in Table~\ref{table:correlationshift}. 
In Section~\ref{sec:ablation}, we discuss a special property of CMNST and propose a modified version of W2D which can achieve a significant gain on CMNIST.

We observe that W2D is the only algorithm that can achieve consistently better performance than ERM on both types of distribution shifts. Specifically, W2D is among the top three in both the datasets dominated by diversity shift as well as the datasets dominated by correlation shift. This comprehensive evaluation supports the view that W2D could serve as a simple heuristic to replace existing training approaches for real-world applications because real-world data 
have
both kinds of distribution shifts.

\begin{table*}
\centering 
\begin{tabular}{l| l l l l | r r}
\hline
Algorithm & PACS & OfficeHome & TerraInc & Camelyon & Average & Ranking score \\
\hline \hline
\textbf{W2D} & $83.4\pm0.3$ & $63.5\pm0.1$ & $44.5\pm0.5$ & $95.2\pm0.3$ & 71.7 & +3\\
RSC~\cite{huang2020self}     & $82.8\pm0.4$ & $62.9\pm0.4$ & $43.6\pm0.5$ & $94.9\pm0.2$ & 71.1 & +2\\
MMD~\cite{li2018domain}     & $81.7\pm0.2$ & $63.8\pm0.1$ & $38.3\pm0.4$ & $94.9\pm0.4$ & 69.7 & +2\\
SagNet~\cite{nam2019reducing}  & $81.6\pm0.4$ & $62.7\pm0.4$ & $42.3\pm0.7$ & $95.0\pm0.2$ & 70.4 & +1\\
\textit{ERM}~\cite{vapnik1999nature} & $81.5\pm0.0$ & $63.3\pm0.2$ & $42.6\pm0.9$ & $94.7\pm0.1$ & 70.5 & 0\\
IGA~\cite{koyama2020out}     & $80.9\pm0.4$ & $63.6\pm0.2$ & $41.3\pm0.8$ & $95.1\pm0.1$ & 70.2 & 0\\
CORAL~\cite{sun2016deep}   & $81.6\pm0.6$ & $63.8\pm0.3$ & $38.3\pm0.7$ & $94.2\pm0.3$ & 69.5 & 0\\
IRM~\cite{arjovsky2019invariant}     & $80.9\pm0.4$ & $63.6\pm0.2$ & $41.3\pm0.8$ & $95.1\pm0.1$ & 70.2 & 0\\
VREx~\cite{krueger2021out}    & $81.8\pm0.4$ & $63.5\pm0.1$ & $40.7\pm0.7$ & $94.1\pm0.3$ & 70.0 & -1\\
GroupDRO~\cite{sagawa2019distributionally}& $80.4\pm0.3$ & $63.2\pm0.2$ & $36.8\pm1.1$ & $95.2\pm0.2$ & 68.9 & -1\\
ERDG~\cite{zhao2020domain}    & $80.5\pm0.5$ & $63.0\pm0.4$ & $41.3\pm1.2$ & $95.5\pm0.2$ & 70.1 & -2\\
DANN~\cite{ganin2016domain}    & $81.1\pm0.4$ & $62.9\pm0.6$ & $39.5\pm0.2$ & $94.9\pm0.0$ & 69.6 & -2\\
MTL~\cite{blanchard2017domain}     & $81.2\pm0.4$ & $62.9\pm0.2$ & $38.9\pm0.6$ & $95.0\pm0.1$ & 69.5 & -2\\
Mixup~\cite{zhang2017mixup}   & $79.8\pm0.6$ & $63.3\pm0.5$ & $39.8\pm0.3$ & $94.6\pm0.3$ & 69.4 & -2\\
ANDMask~\cite{parascandolo2020learning} & $79.5\pm0.0$ & $62.0\pm0.3$ & $39.8\pm1.4$ & $95.3\pm0.1$ & 69.2 & -2\\
ARM~\cite{zhang2020adaptive}     & $81.0\pm0.4$ & $63.2\pm0.2$ & $39.4\pm0.7$ & $93.5\pm0.6$ & 69.3 & -3\\
MLDG~\cite{li2018learning}    & $73.0\pm0.4$ & $52.4\pm0.2$ & $27.4\pm2.0$ & $91.2\pm0.4$ & 61.0 & -4\\
\hline
\end{tabular}
\caption{Performance of domain generalization algorithms on datasets dominated by diversity
shift. W2D achieves better performance than ERM on three datasets with top 1 ranking score.} 
\label{table:diversityshift}
\end{table*}

\begin{table*}
\centering 
\begin{tabular}{l| l l l | r r r}
\hline
Algorithm & CMNIST & NICO & CelebA & Average & Prev score & Ranking score \\
\hline \hline
VREx~\cite{krueger2021out}    & $56.3\pm1.9$ & $71.5\pm2.3$ & $87.3\pm0.2$ & 71.7 & -1 & +1\\
GroupDRO~\cite{sagawa2019distributionally}& $32.5\pm0.2$ & $71.0\pm0.4$ & $87.5\pm1.1$ &  63.7 & -1 & +1\\
\textbf{W2D} &$31.0\pm0.3$ & $71.9\pm1.2$ & $87.7\pm0.4$ &  63.5 & +3 & +1\\
\textit{ERM}~\cite{vapnik1999nature} &$29.9\pm0.9$ & $72.1\pm1.6$ & $87.2\pm0.6$ & 63.1 & 0 & 0\\
IRM~\cite{arjovsky2019invariant}     &$60.2\pm2.4$ & $73.3\pm2.1$ & $85.4\pm1.2$ & 73.0 & -1 & 0\\
ERDG~\cite{zhao2020domain}    &$31.6\pm1.3$ & $72.7\pm1.9$ & $84.5\pm0.2$ & 62.9 & -2 & 0\\
ARM~\cite{zhang2020adaptive}    &$34.6\pm1.8$ & $67.3\pm0.2$ & $86.6\pm0.7$ & 62.8 & -3 & 0\\
MTL~\cite{blanchard2017domain}     &$29.3\pm0.1$ & $70.6\pm0.8$ & $87.0\pm0.7$ & 62.3 & -2 & 0\\
MMD~\cite{li2018domain}     &$50.7\pm0.1$ & $68.9\pm1.2$ & $86.0\pm0.5$ & 68.5 & +2 & -1\\
RSC~\cite{huang2020self}     &$27.6\pm1.8$ & $74.3\pm1.9$ & $85.9\pm0.2$ & 62.6 & +2 & -1\\
Mixup~\cite{zhang2017mixup}   &$28.6\pm1.5$ & $72.5\pm1.5$ & $87.5\pm0.5$ & 62.5 & -2 & -1\\
CORAL~\cite{sun2016deep}   &$30.0\pm0.5$ & $70.8\pm1.0$ & $86.3\pm0.5$ & 62.4 & -1 & -1\\
IGA~\cite{koyama2020out}     &$29.7\pm0.5$ & $71.0\pm0.1$ & $86.2\pm0.7$ & 62.3 & 0 & -1\\
MLDG~\cite{li2018learning}    &$32.7\pm1.1$ & $66.6\pm2.4$ & $85.4\pm1.3$ & 61.6 & -4 & -1\\
SagNet~\cite{nam2019reducing}  &$30.5\pm0.7$ & $69.8\pm0.7$ & $85.8\pm1.4$ & 62.0 & +1 & -2\\
ANDMask~\cite{parascandolo2020learning} &$27.2\pm1.4$ & $71.2\pm0.8$ & $86.2\pm0.2$ & 61.5 & -2 & -2\\
DANN~\cite{ganin2016domain}    &$24.5\pm0.8$ & $69.4\pm1.7$ & $86.0\pm0.4$ & 60.0 & -2 & -3\\

\hline
\end{tabular}
\caption{Performance of domain generalization algorithms on datasets dominated by correlation shift. Prev score indicates the ranking score produced in Table \ref{table:diversityshift}. Although W2D is in third place, the top three methods have the same ranking score, and the gap in averaged accuracy mainly comes from the simplest dataset CMNIST.} 
\label{table:correlationshift}
\end{table*}

\section{Ablation Study}\label{sec:ablation}
There are altogether four hyperparameters for W2D, 
two of them are directly inherited from RSC \cite{huang2020self}: 
\emph{feature dropping percentage}, $\phi$, controls the different dropping percentages to mute feature maps; 
\emph{batch dropping percentage}, $\beta$, controls the different batch size percentages to apply feature dropping. 
There are two new hyperparameters introduced along the sample dimension by W2D:
\emph{worse-case sample percentage}, $\rho$, controls the fraction of the batch size samples with the highest loss used for training;
\emph{whole batch patching percentage}, $\kappa$, controls the percentage of training time trained using the whole batch. 



We use the default hyperparameters from RSC for feature dropping percentage and batch dropping percentage. For selecting the worst-case along the sample dimension, we conduct two ablation studies on possible configurations for it on the standard benchmarks. All results are produced by following Ood-bench's setting. 

Overall, we set the hyperparameter search space of W2D as $\phi \in [0.1, 0.4]$, $ \beta \in [0.1, 0.3]$, $\rho \in [0.1, 0.5]$, $\kappa \in [0.2, 0.4]$.

\subsection{Effect of Worst-case sample percentage $\rho$}
We test W2D with different percentages of worst-case batch samples in Table \ref{table:topp}. For PACS, the fewer the worst-case samples used for training, the higher the test-validation accuracy.
This result suggests that focusing on more hard-to-learn worse-case samples can better push the limit of the model's potential generalization power as indicated by the higher test-validation accuracy.

\begin{table}[h!]
\begin{center}
\begin{tabular}{c |  c |c }
\hline
Percentage & Dataset & Acc(Train-Val/Test-Val) \\
\hline \hline
10 & PACS & 82.4 / 83.7\\
20 & PACS & 83.0 / 83.5\\
33 & PACS & 82.7 / 83.2\\
50 & PACS & 82.7 / 83.1\\
\hline 
\end{tabular}
\caption{Ablation study of Top Worst-case $\rho\%$. We fix other hyperparamters here and only change $\rho$.}
\label{table:topp}
\end{center}
\end{table}

\subsection{Effect of Whole Batch Patching Percentage $\kappa$}
\vspace{-10pt}
In Table \ref{table:batchlastk}, we vary $\kappa$: 0 percent means never training with the whole batch. As we increase $\kappa$, we observe higher training-validation accuracy, but lower testing-validation accuracy. This ablation study demonstrates that whole batch training can 
boost training validation results, while slightly decreasing the model's potential generalization ability at the same time.

\begin{table}[h!]
\begin{center}
\begin{tabular}{c | c |c }
\hline
Percentage & Dataset & Acc(Train-Val/Test-Val) \\
\hline \hline
0 & PACS & 82.2 / 83.7\\
5 & PACS & 82.5 / 83.5\\
10 & PACS & 82.7 / 83.3\\
20 & PACS & 83.0 / 83.3\\
40 & PACS & 82.9 / 83.3\\
\hline 
\end{tabular}
\caption{Ablation study of Whole Batch Training During The Last $\kappa\%$ Epoch. We fix other hyperparameters here and only change $\kappa$.}
\label{table:batchlastk}
\end{center}
\end{table}

\begin{table}[h!]
\begin{center}
\begin{tabular}{c| c | c }
\hline
Methods & Dataset & Acc(Train-Val/Test-Val)  \\
\hline \hline
ERM & PACS & 81.5 / 82.2 \\
feature-dim. & PACS & 82.8 / 83.3 \\
sample-dim. & PACS & 82.2 / 83.5 \\
W2D & PACS & 83.4 / 84.0\\
\hline 
ERM & OfficeHome & 63.3 / 63.5 \\
feature-dim. & OfficeHome & 62.9 / 63.3 \\
sample-dim. & OfficeHome & 63.3 / 63.7 \\
W2D & OfficeHome & 63.5 / 63.8 \\
\hline 
ERM & TerraInc & 42.6 / 43.9 \\
feature-dim. & TerraInc & 43.6 / 44.8 \\
sample-dim. & TerraInc & 42.9 / 45.1 \\
W2D & TerraInc & 44.5 / 46.3 \\
\hline 
ERM & CelebA & 86.3 / 87.2 \\
feature-dim. & CelebA & 86.2 / 85.9 \\
sample-dim. & CelebA & 85.8 / 87.4\\
W2D & CelebA & 86.5 / 87.7 \\
\hline 
\end{tabular}
\caption{Analysis of each dimension of W2D.}
\label{table:Dimensions&TrainTest}
\end{center}
\end{table}

\subsection{Dimensions of Worst-case Training}
\vspace{-5pt}
In Table \ref{table:Dimensions&TrainTest}, we evaluate each component of W2D. Both components (sample dimension and feature dimension) are shown to outperform ERM. We believe each component can be easily plugged into other domain generalization methods and achieve consistent gains.
Also, integrating both components is the best setting (W2D) for most of the diversity shift and correlation shift datasets.


\subsection{Training Validation vs Testing Validation}
\label{sec:trainvstest}
\vspace{-5pt}
For training-domain validation, each training domain is split into training and validation subsets. The models are trained using the training subsets and the final model is chosen as the one that maximizes the accuracy on the union of the validation subsets. Training validation is designed to apply on real-world applications.
For testing-domain validation, the model is selected by maximizing the accuracy on a validation set that follows the distribution of the test domain. Testing validation is used to measure a method's highest potential generalization ability.
In Table \ref{table:Dimensions&TrainTest}, we see that compared to RSC (feature-dim.), W2D tends to obtain bigger improvements when testing-domain validation is used for model selection than training-domain validation. This is mainly because worst-case training along sample dimension can increase the model's potential generalization power.

\subsection{A special property of Colored MNIST}
\label{sec:cmnistproperty}
As mentioned earlier, we evaluate the results in CMNIST using the -90 as testing environment in Table \ref{table:correlationshift} following Ood-Bench\cite{ye2021ood}. In this section, we report the results averaged over three environments (+90, +80 and -90) in CMNIST, 
which is the protocol used in DomainBed
\cite{gulrajani2020search}. 
The study of these results leads us to notice 
a special property of CMNIST in comparison to other methods used. 
Then, this special property leads us to introduce a modified version of W2D that improves over ERM by a clear margin by taking advantage of this property.

Treating the -90 domain as a testing domain is considered to be the most difficult setting because the training and testing domain's distribution are totally flipped. (In comparison, the discussions of the other two testing domains, +90 and +80, are omitted as they are much simpler.)
If we can train on only a small subset of the training samples that share the same distribution as that of testing, the results could be largely improved.  Worst-case training along the sample dimension would be a natural solution for this problem. 
However, we found that a vanilla usage of this method can heavily affect training for this toy dataset.

To solve this problem, we first train a biased classifier at the beginning of training with a few epochs. The biased classifier is then fixed and used as a pre-trained classifier to select the worse-case samples. Specifically, we utilize the worse-case samples selected in each iteration by the biased classifier to train a debiased classifier. Since the distribution flips between the training and testing domain (from +80/+90 to -90), the worse-case samples selected by the pre-trained biased classifier should share a similar distribution with the samples during testing, which leads to surprisingly high performance in Table \ref{table:CMNST}. 

We hope this ablation study can motivate the community to rethink the evaluation method of Colored MNIST. We conjecture a more reasonable protocol is to, rather than only reporting results on -90, evaluate the methods with multiple different distribution domains: \textit{e.g.}, averaged over +/-90, +/-70, +/-50, +/-30, +/-10 domains. 

\begin{table}[h!]
\begin{center}
\begin{tabular}{c |  c |c }
\hline
Method & Dataset & Acc(Train-Val/Test-Val) \\
\hline \hline
ERM\cite{vapnik1999nature} & CMNIST & 51.5 /  58.5\\
GroupDRO\cite{sagawa2019distributionally} & CMNIST &  52.1 / 61.2\\
VREx\cite{krueger2021out} & CMNIST &  51.8 /  56.3\\
ARM\cite{zhang2020adaptive} & CMNIST &  56.2 /  63.2\\
IRM\cite{arjovsky2019invariant} & CMNIST & 52.0 / 70.2 \\
RSC\cite{huang2020self} & CMNIST & 51.7 / 58.5\\
W2D & CMNIST & 51.9 / 59.0 \\
W2D* & CMNIST & \textbf{70.8} / \textbf{72.9}\\
\hline 
\end{tabular}
\caption{The CMNST results are adopted from \cite{gulrajani2020search} and averaged over three domains. * means modified version of W2D.}
\label{table:CMNST}
\end{center}
\end{table}

\section{Discussion}
\label{sec:dis}
\paragraph{Additional Benefit with Stochastic Weight Averaging}
Stochastic Weight Averaging (SWA)~\cite{izmailov2018averaging} is an ensemble technique that finds the solution at the center of a wide flat region of the loss landscape. It performs an equal averaging of the model parameters derived from multiple local minima during the training procedure. SWA was shown to improve the performance in semi-supervised learning and domain adaptation \cite{athiwaratkun2018there}.

In addition to whole batch training, SWA is another effective way to improve the model's generalization at later epoches. 
Intuitively, SWA is able to leverage the worse-case samples from different training stages regardless of whether the samples the model considered worse-case in previously epoches later switches to be easy ones or not.
In Table \ref{table:SWA}, we notice that SWA works especially well with worse-case-based methods. For example, in PACS, W2D obtains a $1.3\%$ improvement from SWA while ERM and feature-dim. (RSC) obtains $0.9\%$ and $0.7\%$ improvement, respectively.

\begin{table}[h!]
\begin{center}
\begin{tabular}{c |  c |c }
\hline
Method & Dataset & Acc(Train-Val/Test-Val) \\
\hline \hline
ERM & PACS & 81.5 / 82.2\\
feature-dim. & PACS & 82.8 / 83.3\\
sample-dim. & PACS & 82.2 / 83.5\\
W2D & PACS & 83.4 / 84.0\\
\hline 
ERM(w SWA) & PACS & 82.5 / 83.0\\
feat-dim.(w SWA) & PACS & 83.5 / 83.7\\ 
sam-dim.(w SWA) & PACS & 83.4 / 83.7\\
W2D(w SWA) & PACS & 84.7 / 84.8\\
\hline 
\end{tabular}
\caption{W2D can further improve the performances if used together with Stochastic Weight Averaging. We apply SWA at the last 25 percent of training time and do not apply whole batch patching here.}
\label{table:SWA}
\end{center}
\vspace{-0.1in}
\end{table}

\vspace{-20pt}
\paragraph{Challenges for our method in DomainBed}

First, methods such as RSC \cite{huang2020self} or the family of DRO methods \cite{sagawa2019distributionally, krueger2021out} are simple heuristic
extensions of ERM along the feature or sample dimension. It seems counter-intuitive that these methods cannot compete with ERM if used properly. 
When closely studying the experimental settings in DomainBed, 
first, we notice the hyperparameter range of RSC goes as high as dropping 50\% of the features, 
and with such a high dropout rate, we find that RSC can barely learn any useful patterns. 
Second, DomainBed changes the default model setting of ResNet50\cite{he2016deep} by adding dropout~\cite{srivastava2014dropout} in the fully connected layers. The highest dropout rate in DomainBed is 50\%, which might be beneficial to other algorithms but could degrade RSC's performance due to the overuse of dropout from both aspects.
For the other dimension, the batchsize range goes as small as 8, which limits the potential of the DRO-family methods to use the hard samples. 

\vspace{-15pt}
\paragraph{Why Ood-bench?}
First, Ood-bench uses a ranking score to reflect a relative degree of robustness against both kinds of distribution shift instead of mean accuracy over different datasets, which is more reasonable. Several algorithms are superior to ERM in the toy case datasets, but they are still vulnerable to the distribution shift from the real data. Thus, using mean accuracy is a less meaningful way to compare these algorithms.
Second, unlike DomainBed, Ood-bench uses a smaller model, ResNet18\cite{he2016deep}, for all algorithms and datasets excluding Colored MNIST. It is known that larger models are usually more robust to distribution shift data and thus their performance may be more easily saturated on small datasets\cite{hendrycks2021many}. Thus, using a smaller base model on which to build on could provide a better testbed for OoD generalization of different algorithms.
Third, the search space of the non-algorithm-specific hyperparameters is carefully designed, such as the learning rate. It allows each algorithm to converge during training at each run.
More importantly, Ood-bench measures each method's generalization ability in a more objective and fair manner: it excludes previous domain generalization techniques that can be plugged into any algorithm, such as dropout~\cite{srivastava2014dropout}. 

\vspace{-15pt}
\paragraph{Limitations}
In the more realistic dataset dominated by correlation shift, such as CelebA and NICO, although W2D achieves the best improvement among all the algorithms, it does not surpasses ERM by a statistically significant margin. 
Despite the high empirical performances, 
there are no sufficient evidence suggesting that 
W2D will be ideal when facing common challenge such as spurious correlations. 
Study to extend W2D to further overcome these challenges is likely expected in the future. 

\section{Conclusion}
\label{sec:con}
\vspace{-5pt}
Inspired by a simple heuristic that 
training with a particular focus on hard-to-learn concepts 
will benefit the learning process, 
in this paper, 
we introduce a training heuristic method that can iteratively force the model to learn the hard-to-learn concepts on both the feature dimension and the sample dimension. 
We name our method W2D following the idea of ``Worst-case along Two Dimensions". 
W2D can be directly applied to almost any model architecture, optimizer, loss, or regularization \textit{etc}. 
After evaluating W2D in OoD-Bench comprehensively, we observe W2D is the only algorithm that can achieve consistently better performance than ERM on both diversity shift and correlation shift.




\vspace{-10pt}
\paragraph{Acknowledgements.}
This work was supported in part by NSF CAREER IIS-2150012 and IIS-2204808. HW was supported by NIH R01GM114311, NIH P30DA035778, and NSF IIS1617583. 

{
\small
\bibliographystyle{ieee_fullname}
\bibliography{ref.bib}
}

\end{document}


\title{Supplementary Material}

\author{
\and
}
\maketitle
\input{latex/commands}

\section{Visualization}
\begin{figure}
    \centering
    \includegraphics[width=0.45\textwidth]{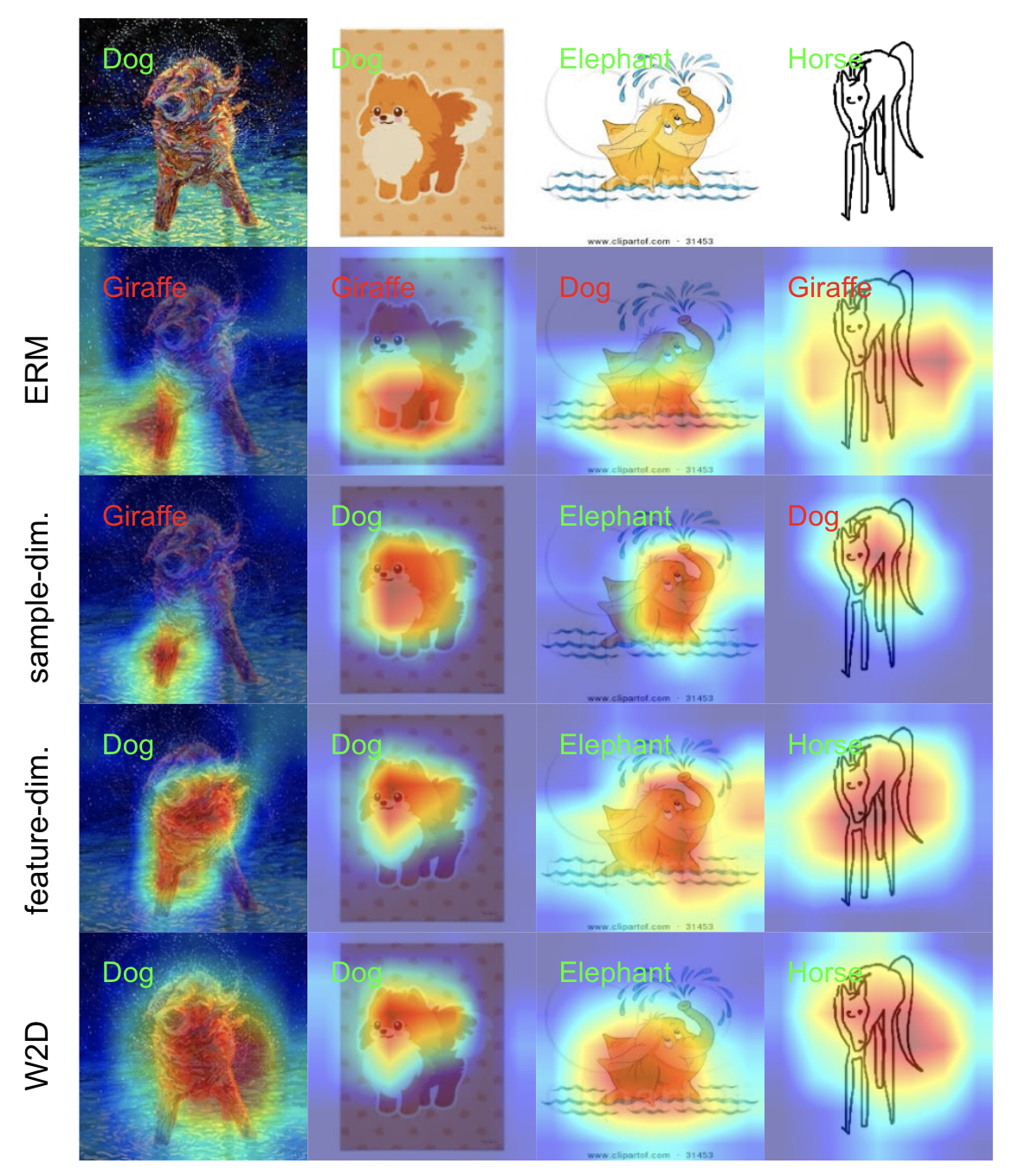}
    \caption{Attention heatmaps of selected algorithms trained and evaluated on PACS\cite{li2017deeper} visualized by class activation map\cite{zhou2016learning}. Green label means correct prediction and red label means wrong prediction.}
    \label{fig:correctpredict}
\end{figure}

In this section, we provide interpretable visualizations of deep neural networks trained by selected algorithms to get a better understanding of the learned representations. In Figure \ref{fig:correctpredict}, we visualize the class attention maps\cite{zhou2016learning} of samples that W2D correctly predicts. Since W2D discards the most predictive representations and forces the model to predict with remaining information, it tends to capture more structural feature information in order to make correct predictions; thus it exhibits broader attention during inference.

\begin{figure}
    \centering
    \includegraphics[width=0.45\textwidth]{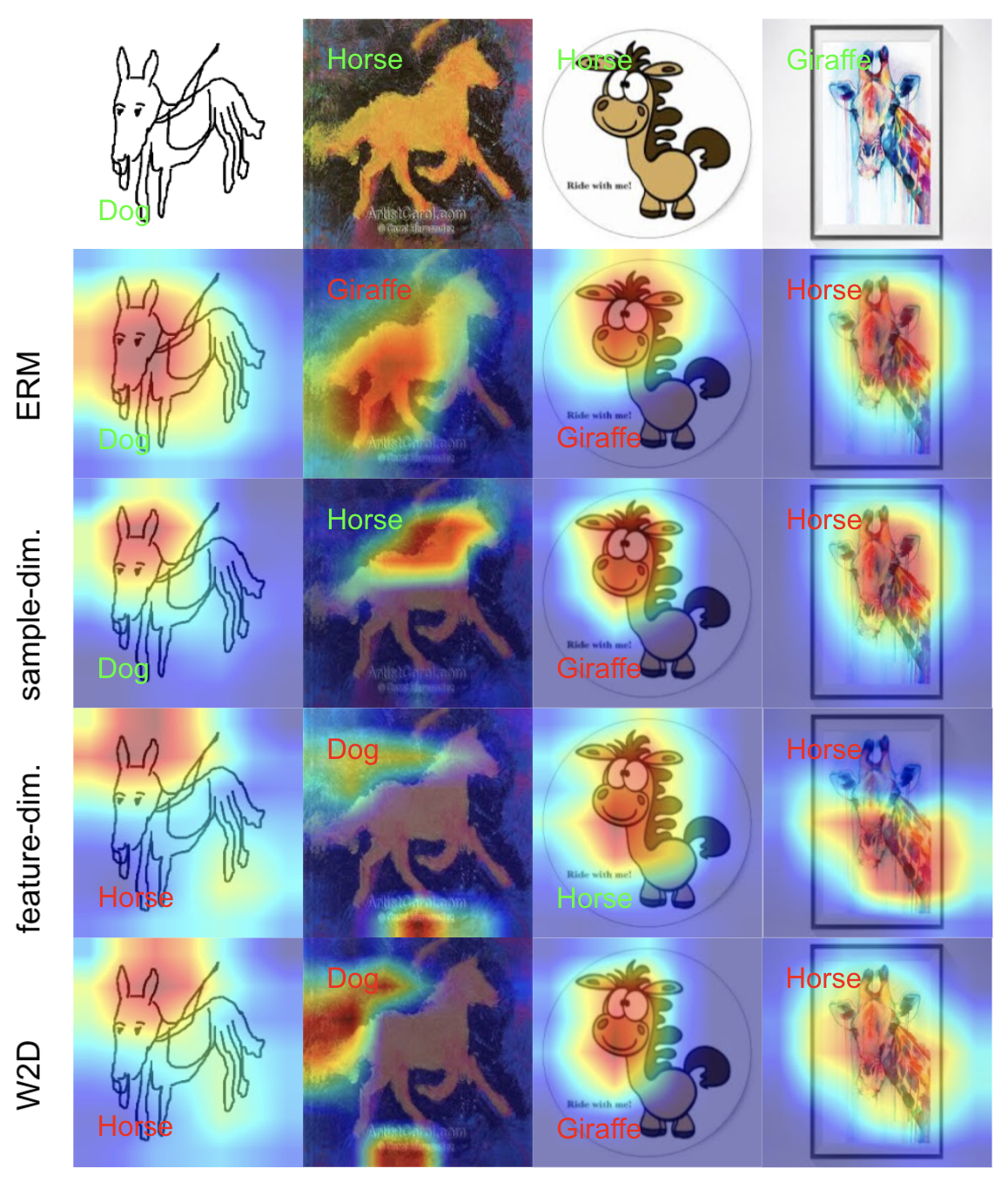}
    \caption{Attention heatmaps of selected algorithms trained and evaluated on PACS\cite{li2017deeper} visualized by class activation map\cite{zhou2016learning}. Green label means correct prediction and red label means wrong prediction.}
    \label{fig:failurepredict}
\end{figure}

On the other hand, 
in Figure \ref{fig:failurepredict}, we visualize the class attention maps\cite{zhou2016learning} of samples that W2D incorrectly predicts. We observe that in the first three columns, W2D fails to make correct predictions while ERM or W2D's components (sample dimension and feature dimension) can predict correctly in these columns. 
Although it appears W2D's performances are degraded over these samples, 
we believe these samples are fairly difficult to predict correctly (even by human) in the first place.

In addition, we visualize the worst-case samples from different domains during training in PACS. We observe that worse-case samples often have rare shapes or textures. Also, the objects in these samples are often partially occluded or viewed from an unusual angle.

\begin{figure}
    \centering
    \includegraphics[width=0.45\textwidth]{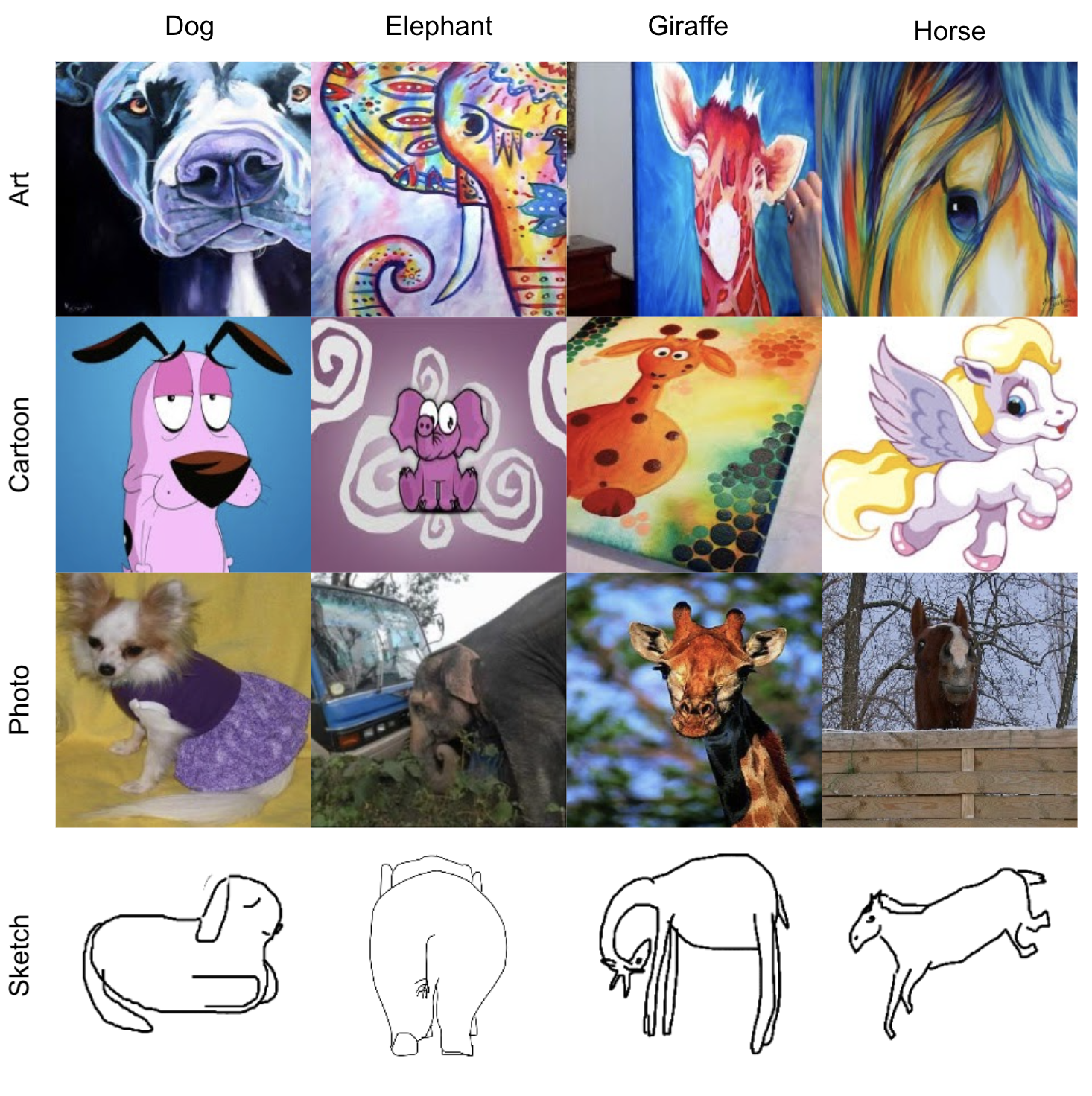}
    \caption{Worst-case samples during training in PACS\cite{li2017deeper}.}
    \label{fig:worst_case}
\end{figure}

\section{Additional Implementation Details}
For network architecture, models trained on CMNIST adopt the two-layer convolution network, while other datasets use ResNet-18 as the backbone following Ood-bench. For hyperparameter search protocol, we use the same as in Ood-bench except for batchsize search space. As we motioned in the discussion section, the batchsize range goes as small as 8 in both Ood-bench and Domainbed, limiting the potential of the DRO-family methods to take advantage of the hard samples. To avoid this issue, we increase the minimum batchsize to 16 in the implementation. 

\section{Additional Empirical Results}
Recall that we evaluate the results in CMNIST using the -90 as testing environment in Table 2 following Ood-Bench\cite{ye2021ood}. In this section, we report the results averaged over three environments (+90, +80 and -90) in CMNIST, which is the protocol used in DomainBed\cite{gulrajani2020search}. The  choice  of  the settings does not affects our ranking score. W2D is still among the top three the datasets dominated by correlation shift.

\begin{table*}
\centering 
\begin{tabular}{l| l  l l | r r r}
\hline
Algorithm & CMNIST & NICO & CelebA & Average & Prev score & Ranking score \\
\hline \hline
GroupDRO&$61.2\pm0.6$ & $71.8\pm0.8$ & $87.5\pm1.1$ &  73.5 & -1 & +1\\
\textbf{W2D}    &$59.1\pm0.3$ & $71.6\pm0.9$ & $87.7\pm0.4$ &  72.8 & +3 & +1\\
ERM &$58.5\pm0.3$ & $71.4\pm1.3$ & $87.2\pm0.2$ & 72.3 & 0 & 0\\
ERDG    &$59.2\pm0.7$ & $70.6\pm1.3$ & $84.5\pm0.2$ & 71.4 & -2 & 0\\
ARM     &$63.2\pm0.1$ & $63.9\pm1.8$ & $86.6\pm0.7$ & 71.2 & -3 & 0\\
IRM     &$70.2\pm0.2$ & $67.6\pm1.4$ & $85.4\pm1.2$ & 74.4 & -1 & -1\\
MMD     &$63.4\pm0.7$ & $68.3\pm1.0$ & $86.0\pm0.5$ & 72.5 & +2 & -1\\
ANDMask &$58.3\pm0.4$ & $72.2\pm1.2$ & $86.2\pm0.2$ & 72.2 & -2 & -1\\
IGA     &$58.7\pm0.5$ & $70.5\pm1.2$ & $86.2\pm0.7$ & 71.8 & 0 & -1\\
MTL     &$57.6\pm0.3$ & $70.2\pm0.6$ & $87.0\pm0.7$ & 71.6 & -2 & -1\\
VREx    &$56.3\pm1.9$ & $71.0\pm1.3$ & $87.3\pm0.2$ & 71.5 & -1 & -1\\
Mixup   &$58.4\pm0.2$ & $66.6\pm0.9$ & $87.5\pm0.5$ & 70.8 & -2 & -1\\
RSC     &$58.5\pm0.2$ & $69.7\pm0.9$ & $85.9\pm0.2$ & 71.4 & +2 & -2\\
SagNet  &$58.2\pm0.3$ & $69.3\pm1.0$ & $85.8\pm1.4$ & 71.1 & +1 & -2\\
DANN    &$58.3\pm0.1$ & $68.6\pm1.1$ & $86.0\pm0.4$ & 71.0 & -2 & -2\\
MLDG    &$58.4\pm0.2$ & $51.6\pm6.1$ & $85.4\pm1.3$ & 65.1 & -4 & -2\\
CORAL   &$57.6\pm0.5$ & $68.3\pm1.4$ & $86.3\pm0.5$ & 70.7 & -1 & -3\\
\hline
\end{tabular}
\caption{Performance of domain generalization algorithms on datasets dominated by correlation shift. Note, the CMNIST results here are adopted from DomainBed\cite{gulrajani2020search}.} 
\label{table:correlationshift_cmnist_version}
\end{table*}

{
\small
\bibliographystyle{ieee_fullname}
\bibliography{ref.bib}
}